\documentclass[10pt,twocolumn,letterpaper]{article}

\usepackage[pagenumbers]{cvpr} 

\usepackage{graphicx}
\usepackage{amsmath}
\usepackage{amssymb}
\usepackage{booktabs}

\usepackage{pifont}
\newcommand{\cmark}{\ding{51}}%

\usepackage{times}
\usepackage{epsfig}
\usepackage{graphicx}
\usepackage{amsmath}
\usepackage{amssymb}
\usepackage{float}
\usepackage{times}
\usepackage{epsfig}
\usepackage{booktabs}
\usepackage{multirow}
\usepackage[table,xcdraw]{xcolor}
\usepackage[accsupp]{axessibility} 
\usepackage{capt-of}
\usepackage[ruled,vlined]{algorithm2e}
\usepackage{multirow}
\usepackage{comment}

\newcommand{\mysubsubsection}[1]{\vspace{0.1cm} \noindent {\bf #1}:}

\definecolor{darkgreen}{rgb}{0.1, 0.7, 0.1}
\definecolor{darkyellow}{rgb}{0.9, 0.7, 0.01}

\usepackage[pagebackref,breaklinks,colorlinks]{hyperref}

\usepackage[capitalize]{cleveref}
\crefname{section}{Sec.}{Secs.}
\Crefname{section}{Section}{Sections}
\Crefname{table}{Table}{Tables}
\crefname{table}{Tab.}{Tabs.}

\def\cvprPaperID{} 
\def\confName{}
\def\confYear{}

\begin{document}

\title{HouseDiffusion: Vector Floorplan Generation via a Diffusion Model \\ with Discrete and Continuous Denoising}

\author{Mohammad Amin Shabani, Sepidehsadat Hosseini, Yasutaka Furukawa\\
Simon Fraser University\\
{\tt\small \{mshabani, sepidh, furukawa\}@sfu.ca}
}


\twocolumn[{
 \maketitle
 \vspace{-2em}
 \centerline{
 \includegraphics[width=\textwidth]{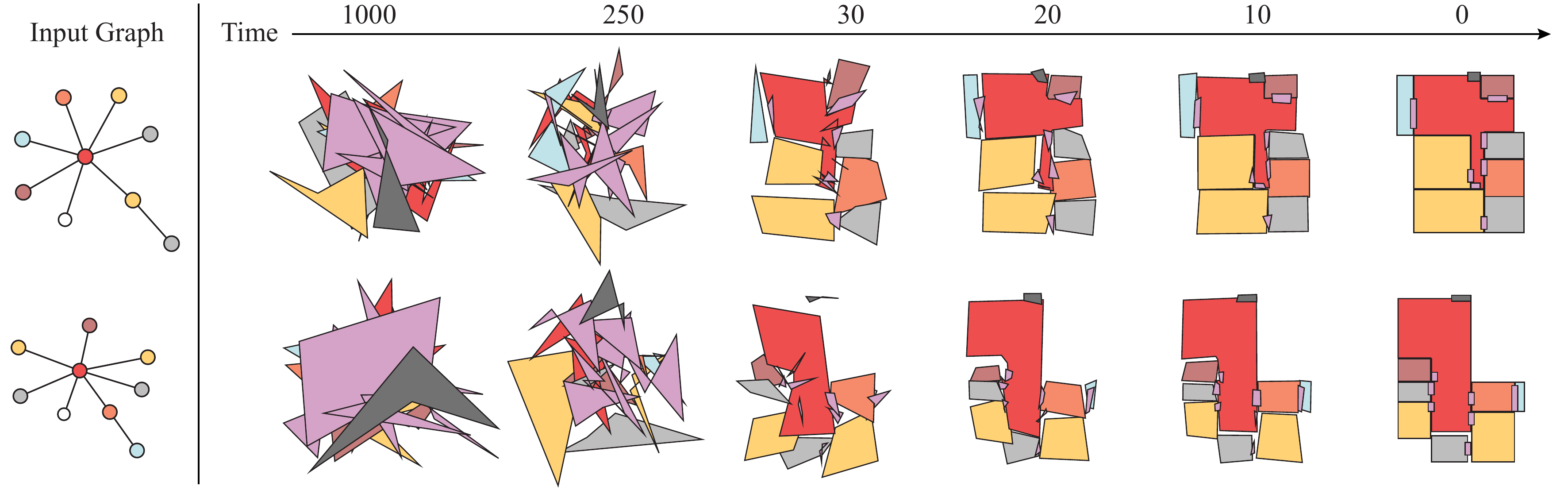} 
}
\captionof{figure}{
Given a bubble diagram as the input constraint, HouseDiffusion directly generates a vector floorplan by initializing the room/door coordinates with  Gaussian noise and iteratively denoising them.
Qualitative and quantitative evaluations demonstrate that HouseDiffusion significantly outperforms the current state-of-the-art with large margins.
%
%
}
\label{fig:teaser}
\vspace{1em}
}]

\begin{abstract}
The paper presents a novel approach for vector-floorplan generation via a diffusion model, which denoises 2D coordinates of room/door corners with two inference objectives: 1) a single-step noise as the continuous quantity to precisely invert the continuous forward process; and 2) the final 2D coordinate as the discrete quantity to establish geometric incident relationships such as parallelism, orthogonality, and corner-sharing.
Our task is graph-conditioned floorplan generation, a common workflow in floorplan design. We represent a floorplan as 1D polygonal loops, each of which corresponds to a room or a door.
%
Our diffusion model employs a Transformer architecture at the core, which controls the attention masks based on the input graph-constraint and directly generates vector-graphics floorplans via a discrete and continuous denoising process.
We have evaluated our approach on RPLAN dataset.
The proposed approach makes significant improvements in all the metrics against the state-of-the-art with significant margins, while being capable of generating non-Manhattan structures and controlling the exact number of corners per room. A project website with supplementary video and document is here \href{https://aminshabani.github.io/housediffusion}{https://aminshabani.github.io/housediffusion}.
\end{abstract}


\section{Introduction}
\label{sec:introduction}

Automated floorplan generation made tremendous progress in the last few years.
While not being fully autonomous yet, the state-of-the-art techniques
help architects to explore the space of possible designs quickly~\cite{housegan,houseganpp}.
90\% of buildings do not have dedicated architects for floorplan design due to their cost in North America. This technology will make the work of professional architects affordable to more house buyers.

Despite recent progress, state-of-the-art floorplan generative models produce samples that are incompatible with the input constraint, lack in variations, or do not look like floorplans~\cite{houseganpp}. The issue is the raster geometry analysis via convolutions, where a room is represented as a binary image. The raster analysis is good at local shape refinement but lacks in global reasoning, and requires non-trivial post-processing for vectorization~\cite{floor-sp,houseganpp}.
On the other hand, direct generation of vector floorplans is not trivial either.
Different from the generation of images or natural languages, structured geometry exhibit precise incident relationships among architectural components (e.g., doors and rooms). For example, a wall is usually axis-aligned, where the coordinate values of adjacent corners are exactly equal. A wall might be shared with adjacent rooms further.
Direct regression of 2D coordinates would never achieve these relationships. 
One could use a discrete representation such as one hot encoding over possible coordinate values with classification, but this causes a label imbalance (i.e., most values are 0 in the encoding) and fails the network training.

This paper presents a novel approach for graph-constrained floorplan generation that directly generates a vector-graphics floorplan (i.e., without any post-processing), handles non-Manhattan architectures, and makes significant improvements on all the metrics. Concretely, a bubble-diagram is given as a graph, whose nodes are rooms and edges are the door-connections. We represent a floorplan as a set of 1D polygonal loops, each of which corresponds to a room or a door, then generate 2D coordinates of room/door corners (See Fig.~\ref{fig:teaser}).
%
The key idea is the use of a Diffusion Model (DM) with a careful design in the 
denoising targets.
Our approach infers 1) a single-step noise amount as a continuous quantity to precisely invert the continuous forward process; and 2) the final 2D coordinate as the discrete quantity to establish incident relationships. The discrete representation after the denoising iterations is the final floorplan model.

Qualitative and quantitative evaluations show that the proposed system outperforms the existing state-of-the-art, House-GAN++~\cite{houseganpp}, with significant margins, while being end-to-end 
and capable of generating non-Manhattan floorplans with exact control on the number of corners per room.
%
We will share all our code and models.
\section{Related Work}

\mysubsubsection{Floorplan generation}
Generation of 3D buildings and floorplans has been an active area of research from a pre-deep learning era~\cite{bao2013generating, hendrikx2013procedural, ma2014game, muller2006procedural, peng2014computing}.
The research area has further flourished with the emergence of deep learning.
Nauata \etal~\cite{housegan} proposed House-GAN as a graph constrained floorplan generative model via Generative Adversarial Network~\cite{goodfellow2020generative}. House-GAN generates segmentation masks of different rooms and combines them to a single floorplan. The authors further improved the quality of the generation by House-GAN++~\cite{houseganpp}, which iteratively refines a layout.
%
Given the boundary of a floorplan, Upadhyay \etal~\cite{upadhyay2022flnet} used the embedded input boundary as an additional input feature to predict a floorplan. Hu \etal~\cite{hu2020graph2plan} proposed Graph2Plan that retrieves a graph layout from a dataset and generates room bounding boxes as well as a floorplan in an ad-hoc way.
%
Sun \etal~\cite{sun2022wallplan} proposed to iteratively generate connectivity graphs of rooms and a floorplan semantic segmentation mask.
%
Given a set of room types and their area sizes as the constraint, Luo and Huang~\cite{luo2022floorplangan} proposed a vector generator and a raster discriminator to train a GAN model using differential rendering. Although their method generates vector floorplans directly, it is limited to rectangular shapes.
Along with the adjacency graph as the input, Yin \etal~\cite{bisht2022transforming} use graph-theoretic and linear optimization techniques to generate floorplans.
%
Our paper also tackles a graph-constrained floorplan generation with a bubble diagram as the constraint~\cite{houseganpp}.
The key difference is that HouseDiffusion processes a vector geometry representation from start to finish, and hence, directly generating vector floorplan samples.


\mysubsubsection{Diffusion models}
Deep generative models have seen great success in broader domains~\cite{goodfellow2020generative, oord2016wavenet, wang2018pixel2mesh, qi2017pointnet, lim2017enhanced}, where a Diffusion Model (DM)~\cite{sohl2015deep,yang2022diffusion,cao2022survey} is an emerging technique.
Ho \etal~\cite{ho2020denoising} used a DM to boost image generation quality.
Dhariwal and Nichol~\cite{nichol2021improved} made improvements by proposing a new noise schedule and learning the variances of the reverse process.
%
The same authors made further improvements by novel architecture and classifier guidance~\cite{dhariwal2021diffusionbeats}.
DMs have been adapted to many other tasks 
such as Natural Language Processing~\cite{li2022diffusion}, Image Captioning~\cite{chen2022analog}, Time-Series Forecasting~\cite{tashiro2021csdi}, Text-to-Speech~\cite{kim2022guided, koizumi2022specgrad}, and finally Text-to-Image as seen in the great success of 
DALL-E 2~\cite{ramesh2022hierarchical} and Imagen~\cite{saharia2022photorealistic}.

Molecular Conformation Generation~\cite{hoogeboom2022equivariant, luo2021predicting, xu2022geodiff, jing2022torsional} and 3D shape generation~\cite{luo2021diffusion, lyu2021conditional, luo2021score, zhou20213d} are probably the closest to our task.
%
What makes our task unique and challenging is the precise geometric incident relationships, such as parallelism, orthogonality, and corner-sharing among different components, which continuous coordinate regression would never achieve.
%
In this regard, several works use discrete state space~\cite{austin2021structured, campbell2022continuous, hoogeboom2021argmax} or learn an embedding of discrete data~\cite{li2022diffusion, chen2022analog} in the DM formulation.
However, we found that these pure discrete representations do not train well, probably because the diffusion process is continuous in nature.
In contrast, our formulation simultaneously infers a single-step noise as the continuous quantity and the final 2D coordinate as the discrete quantity, achieving superior generation capabilities (See Sect.~\ref{sec:ablation_studies} for more analysis). To our knowledge, our work is the first in using DMs to generate structured geometry.
\section{Preliminary}
\label{sec:preliminary}

Diffusion models (DMs) denoise a Gaussian noise $x_T$ towards a data sample $x_0$ in $T$ steps, whose training consists of the forward and the reverse processes.
The forward process takes a data sample $x_0$ and generates a noisy sample $x_t$ at time step $t$ by sampling a Gaussian noise $\epsilon \sim \mathcal{N}(0,\textbf{I})$:
\begin{equation}
    x_t = \sqrt{\gamma_t}x_0 + \sqrt{(1-\gamma_t)}\epsilon. \label{eq:forward}
\end{equation}
$\gamma_t$ is a noise schedule that gradually changes from 1 to 0.
%
The reverse process starts from a pure Gaussian noise $x_T \sim \mathcal{N}(0,1)$ and learns to denoise a sample step by step until reaching $x_0$, where the denoising process takes $x_t$ and estimates $x_{t-1}$ by inferring $x_{t-1}$, $\epsilon$, or $x_0$~\cite{ho2020denoising}.

\section{HouseDiffusion}
\begin{figure*}[tb]
    \centering
    \includegraphics[width=\textwidth]{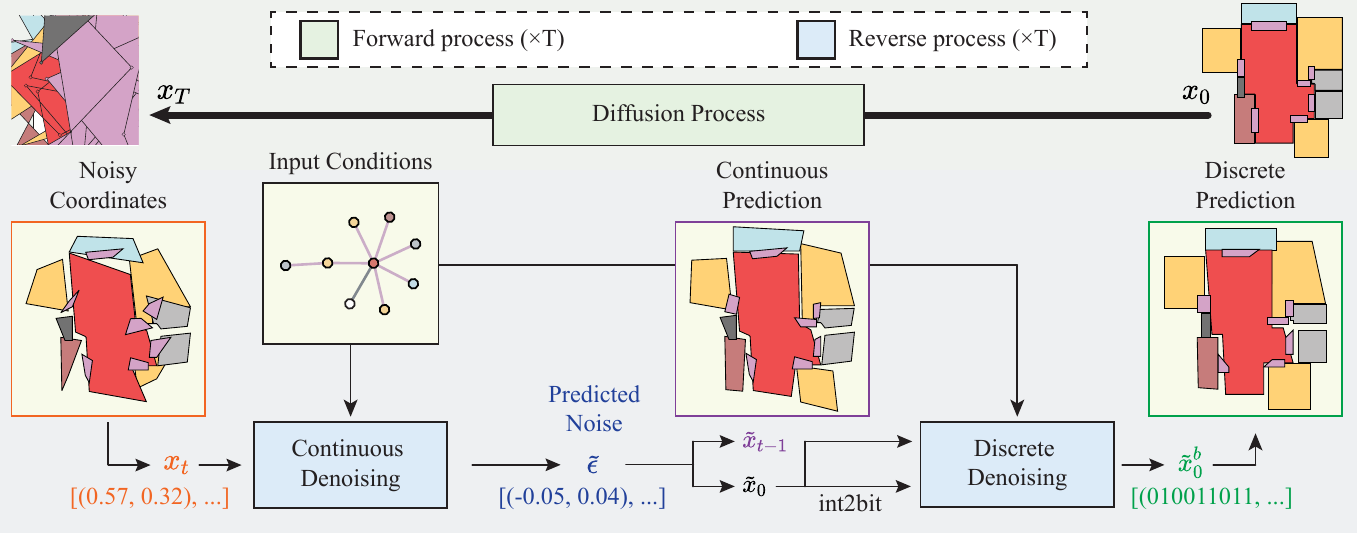}
    \caption{
    The forward process takes the ground-truth floorplan $x_0$ and adds a Gaussian noise to create a noisy floorplan sample $x_t$.
    The reverse process takes 
    a noisy floorplan at time $t$ with a bubble diagram as the condition.
    The process infers the corresponding noise $\Tilde{\epsilon}$ and $\Tilde{x}_0$ in the continuous and the discrete (binary) representations, respectively.
    }
    \label{fig:method} 
\end{figure*}

HouseDiffusion solves a graph-constrained floorplan generation problem (See Fig.~\ref{fig:teaser}). The constraint is a bubble diagram, whose nodes are rooms and edges are door connections. A room node is associated with a room type.~\footnote{Room types are ``Kitchen", ``Living-room", ``Bedroom", ``Dining-room", ``Bathroom", ``Study-room", ``Balcony", ``Entrance", ``Storage", and ``Unknown". Door types are ``Interior door" and ``Front door".}

The choice of a data representation is crucial for the success of any generative model. A floorplan is a graph, whose general representation is a set of room-corners and their connections as an adjacency matrix. However, this is not an easy representation to generate with the bubble diagram constraint.
Instead, we represent a floorplan as a set of 1D polygonal loops
one for each room/door.
The challenge is to ensure geometric consistencies among the loops.
For example, room corners and walls must be shared exactly without gaps or overlaps.
The section explains our solution, namely, the floorplan representation and the network architecture.




\subsection{Floorplan Representation}
\label{sec:polygon_representation}

%
%


Let $P=\{P_1, P_2, ..., P_N\}$ denote the set of polygonal loops for each room/door to be generated.
Each loop $P_i$ is defined by a sequence of corners with 2D coordinates:
\begin{equation}
P_i = \{C_{i,1}, C_{i,2}, ..., C_{i,N_i}|C_{i,j}\in \mathcal{R}^2\}.
\end{equation}
%
$N_i$ denotes the number of corners in $P_i$, which needs to be specified or generated.
A common approach is to set the greatest possible number and let the network decide how many corners to use via output flags.
However, this significantly increases the representation size and makes training harder and inefficient.
Instead, we construct a histogram of the number of corners for each room/door type from the training samples, then probabilistically pick $N_i$ where the probabilities are proportional to their histogram counts. This heuristic works well in practice. A user could also directly specify the number of corners to control the room shape complexity.




Coordinate values $\{C_{i,j}\}$ are integers in the range of $[0, 255]$. In the forward process, we affinely map the range to $[-1, 1]$ and treat as continuous values to be mixed with the Gaussian noise $\mathcal{N}(0,1)$.
In the reverse process, we also represent a coordinate value as a discrete integer in the binary representation, that is, as 8 binary numbers.

\subsection{HouseDiffusion Architecture}
\label{sec:network_architecture}

HouseDiffusion is a diffusion model based architecture.
The forward process follows (\ref{eq:forward}), where $\gamma_t$ is a standard cosine noise schedule~\cite{nichol2021improved}. The reverse process is a Transformer~\cite{vaswani2017attention} based neural network, which takes the floorplan representation at time $t$ and infers the representation at time $t-1$ (See Fig.~\ref{fig:method}). We append a superscript $t$ to denote the floorplan sample at time $t$ (\eg, $P^t_i$ or $C^t_{i,j}$).


\mysubsubsection{Feature embedding}
Given a floorplan sample $\{P^t\}$ at time $t$, every room/door corner $C^t_{i,j}$ becomes a node in the Transformer architecture with a $d (=512)$ dimensional embedding vector $\hat{C}^t_{i,j}$.
We initialize the embedding as
%
\begin{equation}
    \hat{C}^t_{i,j} \leftarrow \mbox{Linear}( [\mbox{AU}(C^t_{i,j}), R_i, \mathbf{1}(i), \mathbf{1}(j), t]) \label{eq:embedding}
\end{equation}
$\mbox{AU}$ augments the corner coordinate by 1) uniformly sampling $L (=8)$ points along the wall to the next corner; and 2) concatenating the sampled point coordinates. The augmentation helps to reason incident relationships along the walls.
%
$R_i$ is a 25D room-type one hot vector. $\mathbf{1}(\cdot)$ denotes a 32D one-hot vector for a room index $i$ and a corner index $j$.
$t$ is a scalar.
A linear layer converts the embedding to a 512D vector.



\begin{figure}[tb]
    \centering
    \includegraphics[width=\linewidth]{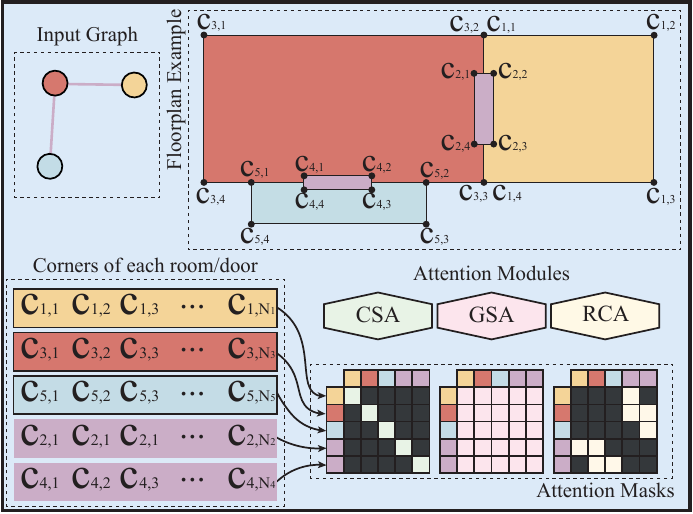}
    \caption{Given the input bubble diagram, our model benefits from three attention modules explicitly processing different levels of relations between coordinates. The figure shows how a group of coordinates share information with other groups in each attention module. The black cells represent the masked-out attentions.}
    \label{fig:model} 
\end{figure}

\mysubsubsection{Continuous denoising}
Embedding vectors $\{\hat{C}^t_{i,j}\}$ will go through attention layers with structured masking (See Fig.~\ref{fig:model}).
There are three types of attentions in our attention layer:
%
1) Component-wise Self Attention (CSA), limiting attentions among nodes in the same room or door
2) Global Self Attention (GSA), a standard self-attention between every pair of corners across all rooms;
and 3) Relational Cross Attention (RCA), limiting attentions to from-room-to-door or from-door-to-room that are connected in the constraint graph.
%
We learn three sets of key/query/value matrices in each of the attention layer per head, while we use four multi-heads.
The results of the three attentions are summed, followed by a standard Add \& Norm layer. The continuous denoising repeats the block of this attention and Add \& Norm layers four times. At the end, a single linear layer infers noise $\epsilon_{\theta}(C_{i,j}, t)$ at each node.

\mysubsubsection{Discrete denoising}
%
Geometric incident relationships (e.g., colinearity, orthogonality, or corner sharing) rarely emerge
through coordinate regression. For example, two coordinates almost never become the same by regression.
Our approach infers coordinates in a discrete form.
Concretely, after obtaining ${C}^{t-1}_{i,j}$ and ${C}^{0}_{i,j}$ from the continuous denoising process, we affinely map $C^{0}_{i,j}$ back to the range $[0, 255]$, apply rounding, and use an ``int2bit" function~\cite{chen2022analog} to convert to a binary representation, that is, 8-dimensional binary vector.
%
We use a similar formula as formula (\ref{eq:embedding}) to obtain a 512D embedding vector for each corner:
\begin{equation}
    \mbox{Linear}( [\mbox{AU}(C^0_{i,j}), \mbox{int2bit}(C^0_{i,j}), R_i, \mathbf{1}(i), \mathbf{1}(j), t]) \label{eq:embedding2}
\end{equation}
We pass ${C}^{0}_{i,j}$ as the input to the augmentation and we pass the binary representation obtained by int2bit function along with the conditions to help the network by providing the initial binary representation. We repeat two blocks of attention with structured masking, followed by a linear layer to produce an 8-dimensional vector as $C^0_{i,j}$. During testing, we binary threshold $C^0_{i,j}$ and obtain the integer coordinate. During training, we directly use the values without thresholding for a loss function.


\mysubsubsection{Loss functions}
We train our model end-to-end with the simple L2-norm regression loss by Ho \etal~\cite{ho2020denoising} on both continuous and discrete regressions with the same weight. Concretely, $\epsilon_{\theta}(C_{i,j}, t)$ is compared with the ground-truth noise $\epsilon$ from the forward process. $C^0_{i,j}$ is compared with the ground-truth corner coordinate in binary representation.
The inference of $C^0_{i,j}$ becomes accurate only near the end of the denoising process. Therefore, the discrete branch is used during training only when $t < 20$. At testing, we use the denoised integer coordinates from the discrete branch to pass to the next iteration only when $t < 32$.

\section{Experiments}

\begin{figure*}[p]
    \centering
    \includegraphics[width=1.01\textwidth]{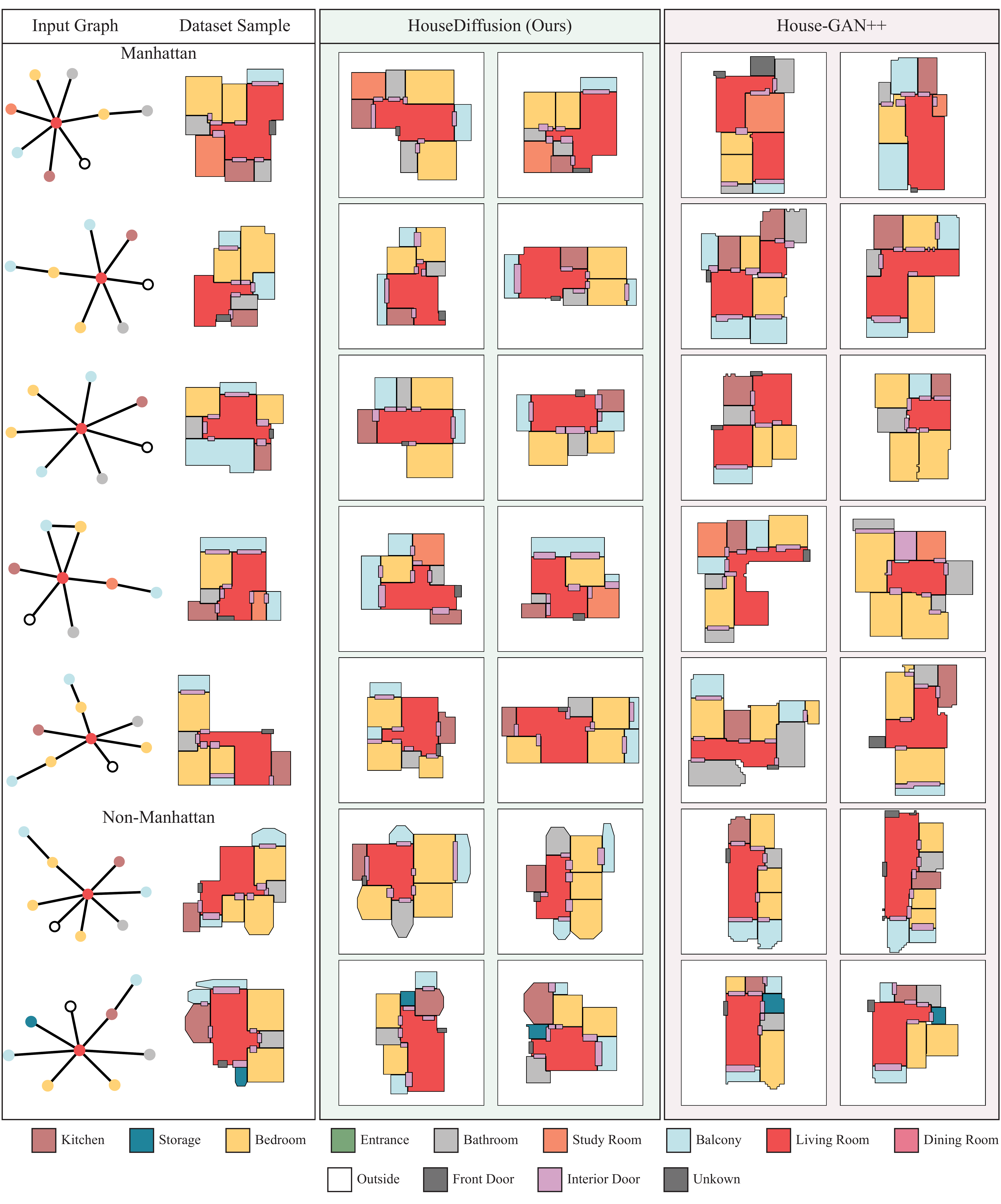}
    \caption{Generated floorplan samples against House-GAN++~\cite{houseganpp}. See supplementary for more examples. Our results look more diverse and higher quality, where the major issue of House-GAN++ is duplicate or missing rooms, ignoring the input constraint.
%
    }
    \label{fig:main_qualitative_results}
\end{figure*}

\begin{figure*}[tb]
    \centering
    \includegraphics[width=\textwidth]{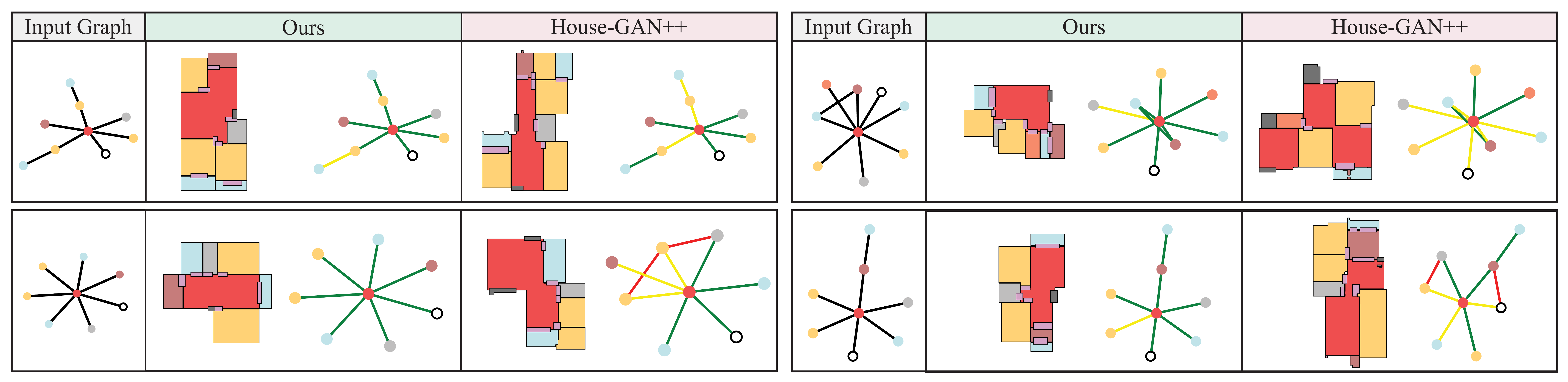}
    \caption{HouseDiffusion results are more compatible with the input bubble diagram by ensuring the generation of a single polygonal loop for each room/door.
    %
    {\color{darkgreen} Green }, {\color{darkyellow} Yellow }, and {\color{red} Red} indicate correct, missing, and extraneous connections, respectively.}
    \label{fig:compatibility_comparison}
\end{figure*}

We use PyTorch to implement the proposed approach based on a public implementation of Guided-Diffusion~\cite{dhariwal2021diffusionbeats}.~\footnote{\href{https://github.com/openai/guided-diffusion}{https://github.com/openai/guided-diffusion}}
Adam~\cite{kingma2014adam} is the optimizer with decoupled weight decay~\cite{loshchilov2018decoupled} for 250k steps with batch-size of 512 on a single NVIDIA RTX 6000. An initial learning rate is 1e-3.
We divide the learning rate by 10 after every 100k steps.
We set the number of diffusion steps to 1000 (unless otherwise noted) and uniformly sample $t$ during training.


We compare against other graph-constrained floorplan generative models (House-GAN++~\cite{houseganpp} and House-GAN~\cite{housegan}) and scene-graph constrained image generative models (Ashual \etal~\cite{ashual2019specifying}, and Johnson \etal~\cite{johnson2018image}). House-GAN++ is the current state-of-the-art for the task.



We use 60,000 vector floorplans from RPLAN~\cite{rplan} dataset,
and the same pre-processing steps as in House-GAN++, where floorplan images have a resolution of $256\times 256$.
We divide the floorplan samples into four groups based on the number of rooms (i.e., 5, 6, 7, or 8 rooms). For generating floorplans in each group, we exclude samples in the group from the training so that methods cannot simply memorize samples.
Following House-GAN++, the same three metrics are used for evaluations: Diversity, Compatibility, and Realism. Diversity is the Frechet Inception Distance (FID)~\cite{fid}.
Compatibility is the modified Graph Edit Distance~\cite{ged} between the input bubble diagram and the one reconstructed from the generated floorplan. Realism is based on user studies (See Section~\ref{sec:main_results} for more details).

To verify the capability of non-Manhattan floorplan generation,
we create a new benchmark ``Non-Manhattan-RPLAN'' based on RPLAN by randomly adding two corners to an outer wall.
See supplementary for more details.





\subsection{Quantitative Evaluations}
\label{sec:main_results}
Table~\ref{tab:main_results} shows the main quantitative evaluations.
For existing methods, we copy the numbers reported in the House-GAN++ paper~\cite{houseganpp}. For the Realism, their paper shows the average score against all the other methods, while we use the score against just the ground-truth, because comparisons between weaker baselines do not provide useful information.
Our system, HouseDiffusion, consistently outperforms all the previous methods in all the metrics. Compared to the current state-of-the-art House-GAN++~\cite{houseganpp}, HouseDiffusion makes an average improvement of $67\%$ in diversity and $32\%$ in compatibility.
We also make significant improvements in realism, which we discuss next in detail.
For non-manhattan RPLAN dataset, HouseDiffusion also makes significant improvements in diversity and compatibility, where House-GAN++ cannot handle non-manhattan structures. The realism score is not used, because the non-manhattan structures were added by heuristics and even the ground-truth may not look realistic.

\begin{table*}[!tb]
\caption{Main quantitative results, comparing HouseDiffusion (ours) with the previous methods on the three metrics. In all experiments, our method significantly outperform all the other methods.}
\label{tab:main_results}
\centering
\setlength{\tabcolsep}{3pt}
\small
\begin{tabular}{c|c|c|cccc|cccc}
\toprule
   & \multicolumn{1}{c}{} & \multicolumn{1}{c}{Realism $(\uparrow)$} & \multicolumn{4}{c}{Diversity $(\downarrow)$} & \multicolumn{4}{c}{Compatibility $(\downarrow)$} \\
    \cmidrule(lr){3-3}\cmidrule(lr){4-7}\cmidrule(lr){8-11} Dataset & Model &
    8 & 5 & 6 & 7 & 8 & 5 & 6 & 7 & 8 \\
    \midrule
    \multirow{5}{*}{{RPLAN}} & 
    Ashual \etal~\cite{ashual2019specifying} & -1.00 & 120.6\scalebox{0.8}{$\pm$0.5} & 172.5\scalebox{0.8}{$\pm$0.2} & 162.1\scalebox{0.8}{$\pm$0.4} & 183.0\scalebox{0.8}{$\pm$0.4} & 7.5\scalebox{0.8}{$\pm$0.0} & 9.2\scalebox{0.8}{$\pm$0.0} & 10.0\scalebox{0.8}{$\pm$0.0} & 11.8\scalebox{0.8}{$\pm$0.0}\\
    & Johnson \etal~\cite{johnson2018image} & -1.00 & 167.2\scalebox{0.8}{$\pm$0.3} & 168.4\scalebox{0.8}{$\pm$0.4} & 186.0\scalebox{0.8}{$\pm$0.4} & 186.0\scalebox{0.8}{$\pm$0.4} & 7.7\scalebox{0.8}{$\pm$0.0} & 6.5\scalebox{0.8}{$\pm$0.0} & 10.2\scalebox{0.8}{$\pm$0.0} & 11.3\scalebox{0.8}{$\pm$0.1}\\
    & House-GAN~\cite{housegan} & -0.95 &  37.5\scalebox{0.8}{$\pm$1.1} & 41.0\scalebox{0.8}{$\pm$0.6} & 32.9\scalebox{0.8}{$\pm$1.2} & 66.4\scalebox{0.8}{$\pm$1.7} & 2.5\scalebox{0.8}{$\pm$0.1} & 2.4\scalebox{0.8}{$\pm$0.1} & 3.2\scalebox{0.8}{$\pm$0.0} & 5.3\scalebox{0.8}{$\pm$0.0}\\
    & House-GAN++~\cite{houseganpp} & -0.52 & 30.4\scalebox{0.8}{$\pm$4.4} & 37.6\scalebox{0.8}{$\pm$3.0} & 27.3\scalebox{0.8}{$\pm$4.9} & 32.9\scalebox{0.8}{$\pm$4.9} & 1.9\scalebox{0.8}{$\pm$0.3} & 2.2\scalebox{0.8}{$\pm$0.3} & 2.4\scalebox{0.8}{$\pm$0.3} & 3.9\scalebox{0.8}{$\pm$0.5}\\
    & Ours & \textbf{-0.19} & \textbf{11.2}\scalebox{0.8}{$\pm$0.2} & \textbf{10.3}\scalebox{0.8}{$\pm$0.2} & \textbf{10.4}\scalebox{0.8}{$\pm$0.4} & \textbf{9.5}\scalebox{0.8}{$\pm$0.1} & \textbf{1.5}\scalebox{0.8}{$\pm$0.0} & \textbf{1.2}\scalebox{0.8}{$\pm$0.0} & \textbf{1.7}\scalebox{0.8}{$\pm$0.0} & \textbf{2.5}\scalebox{0.8}{$\pm$0.0}\\
    
    \midrule
    \multirow{2}{*}{{NM-RPLAN}} &
House-GAN++~\cite{houseganpp} & --- & 77.3\scalebox{0.8}{$\pm$0.8} & 60.0\scalebox{0.8}{$\pm$0.7} & 73.8\scalebox{0.8}{$\pm$0.8} & 58.2\scalebox{0.8}{$\pm$1.0} & 1.5\scalebox{0.8}{$\pm$0.0} & 2.9\scalebox{0.8}{$\pm$0.0} & 2.1\scalebox{0.8}{$\pm$0.0} & 3.2\scalebox{0.8}{$\pm$0.0}\\
& Ours & --- & \textbf{12.0}\scalebox{0.8}{$\pm$0.2} & \textbf{11.0}\scalebox{0.8}{$\pm$0.1} & \textbf{10.3}\scalebox{0.8}{$\pm$0.2} & \textbf{10.5}\scalebox{0.8}{$\pm$0.3} & \textbf{1.2}\scalebox{0.8}{$\pm$0.0} & \textbf{1.3}\scalebox{0.8}{$\pm$0.0} & \textbf{1.6}\scalebox{0.8}{$\pm$0.0} & \textbf{2.5}\scalebox{0.8}{$\pm$0.0}\\

    \bottomrule
\end{tabular}
\end{table*}


We follow the same process as in House-GAN++ to obtain the realism scores.
We generate 1000 floorplan samples by each system, present two samples (from different systems) to a participant, and ask to choose ``A is better'', ``B is better'', or ``both are equal''.
A method ``A'' earns +1, -1, or 0 point with the above answers, respectively.
Each pair of systems are evaluated 150 times by 10 participants (i.e., 15 times by each participant).
Table~\ref{tab:main_results} shows the average realism score of each system against the ground-truth. Table~\ref{tab:pairwise comparison} shows the direct comparisons between House-GAN++, HouseDiffusion, and the ground-truth.
%
%
Both tables show that HouseDiffusion achieves significant improvements with a score of $-0.19$ even against the ground-truth, which is very good.
The number implies that participants
choose our result to be as realistic as the Ground-Truth (``both are equal'') for $81\%$ of the time (assuming they either preferred Ground-Truth or chose equal).
Note that House-GAN++ paper reports $-0.18$ for their best configuration against the ground-truth, but this variant (denoted as ``Ours static*'') requires expensive post-processing~\cite{floor-sp} and filters out incompatible samples based on the ground-truth bubble-diagram that account for 90\% of the generated samples. Our results are direct outputs from our network architecture.

\begin{table}[tb]
\caption{Realism scores among House-GAN++, ours, and the ground-truth.
Our method achieves 0.71 against House-GAN++,
indicating that the participants choose ours to be more realistic
$85\%$ of the times (i.e., $0.71 \approx 0.85 - 0.15$).
}
\label{tab:pairwise comparison}
\centering
\setlength{\tabcolsep}{3pt}
\small
\scalebox{1.}{\begin{tabular}{c|c|c|c}
\toprule
    & House-GAN++ & Ours & Ground Truth\\
    \cmidrule(lr){1-1}\cmidrule(lr){2-2} \cmidrule(lr){3-3} \cmidrule(lr){4-4}
    House-GAN++~\cite{houseganpp} & --- & -0.71 & -0.87\\
    \midrule
    Ours & 0.71 & --- & -0.19 \\
    \midrule
    Ground Truth & 0.87 &  0.19 & ---\\
    \bottomrule
\end{tabular}}
\end{table}


\subsection{Qualitative Evaluations}
Figure~\ref{fig:main_qualitative_results} qualitatively compares HouseDiffusion with House-GAN++ by two generated floorplans per bubble-diagram in both Manhattan and Non-Manhattan cases. Please refer to the supplementary for more examples and the supplementary video for the animations of the denoising process.
%
HouseDiffusion consistently generates higher-quality samples. The major issue of Housen-GAN++ is that the system tends to miss or generate too many rooms, ignoring the input bubble-diagram constraint. For example, the top example in Fig.~\ref{fig:main_qualitative_results} should have one living-room (red), two bedrooms (yellow), and one study (orange), but House-GAN++ makes errors in every single sample. The issue is also highlighted in
%
Fig.~\ref{fig:compatibility_comparison}, which visualizes the bubble-diagrams of generated floorplan samples.
HouseDiffusion guarantees exactly the correct set of rooms in the output and produces clean wall structures without artifacts (e.g., jagged boundaries) thanks to the vector representation.

\subsection{Ablation Studies}
\label{sec:ablation_studies}

We conduct a series of ablation studies to verify the effectiveness of our technical contributions and provide an in-depth analysis of our system.

\mysubsubsection{Discrete and continuous denoising} 
To verify the effectiveness of our discrete and continuous denoising scheme, we compare with a recent work AnalogBits~\cite{chen2022analog}, which employs a binary representation to generate discrete numeric values in a diffusion model framework.
We also create HouseDiffusion only with continuous denoising by simply dropping the discrete branch/loss.
Table~\ref{tab:diffusion_domain} demonstrates that the proposed discrete and continuous denoising significantly improves the diversity with a small sacrifice on the compatibility. 
We hypothesize that AnalogBits merely learns the rounding with the binary representation alone when $t$ is small, which reduces the generation quality.
%


%

\begin{table}[tb]
\caption{The effectiveness of our discrete and continuous denoising scheme, in comparison to an existing discrete only representation AnalogBits and our system without the discrete branch/loss that reduces to a standard denoising scheme.
}
\label{tab:diffusion_domain}
\centering
\setlength{\tabcolsep}{3pt}
\small
\scalebox{1.}{\begin{tabular}{c|cc|c|c}
\toprule
   & Cont. & Disc. & Divers. $(\downarrow)$ & Compat. $(\downarrow)$\\
    %
    \cmidrule(lr){1-5}
    AnalogBits~\cite{chen2022analog} & & \cmark & 14.5\scalebox{0.8}{$\pm$0.3} & 6.0\scalebox{0.8}{$\pm$0.03}\\
    \midrule
    Ours w/o disc. & \cmark &  & 38.8\scalebox{0.8}{$\pm$0.9} & 2.2\scalebox{0.8}{$\pm$0.0} \\
    \midrule
    Ours & \cmark & \cmark & 9.5\scalebox{0.8}{$\pm$0.1} & 2.5\scalebox{0.8}{$\pm$0.0} \\
    \bottomrule
\end{tabular}}
\end{table}

\begin{figure}[tb]
    \centering
    \includegraphics[width=\linewidth]{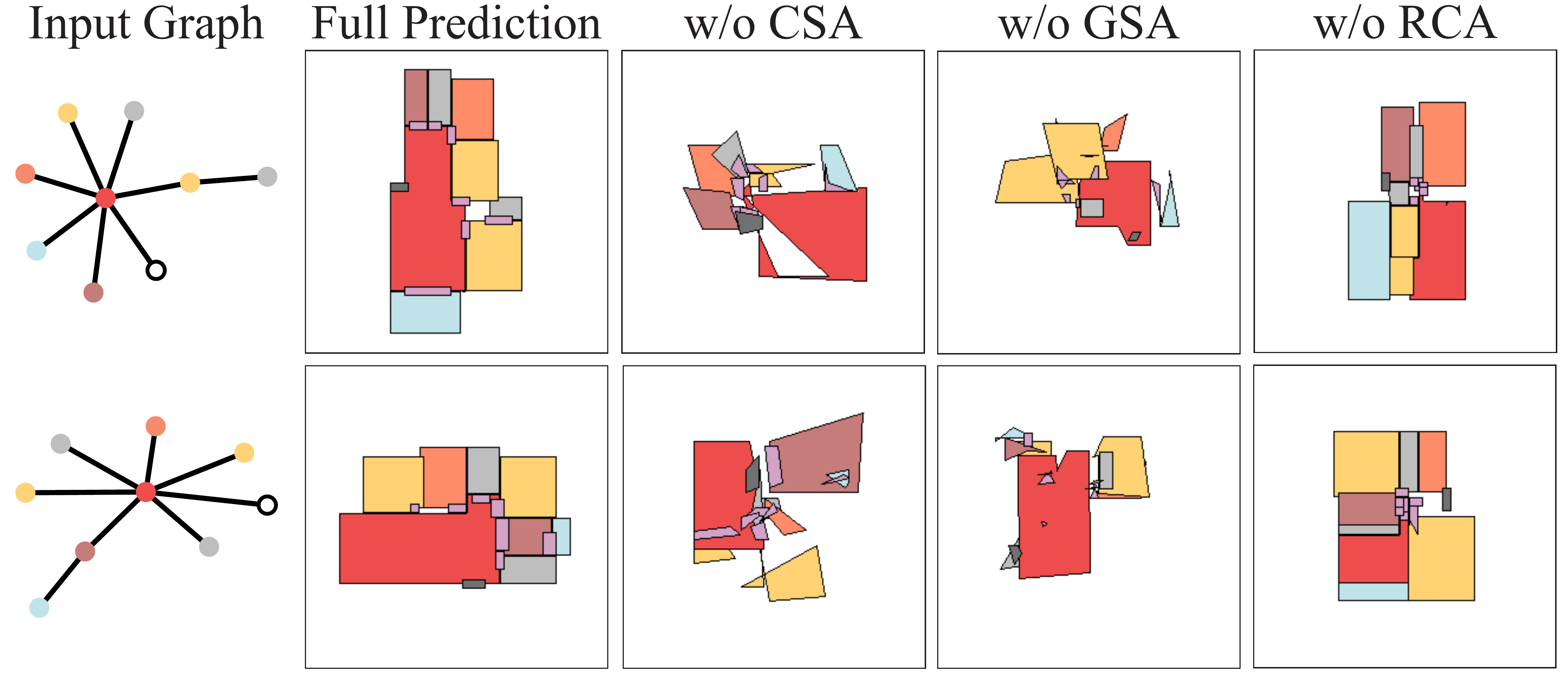}
    \caption{Qualitative evaluation of the three attention mechanisms. The shape quality degrades significantly in the absence of CSA or GSA. Door placements become corrupted in the absence of RCA.
}
    \label{fig:attention_modules}
\end{figure}

\mysubsubsection{Attention modules}
HouseDiffusion employs three different types of attentions.
Table~\ref{tab:different_attentions} and Fig.~\ref{fig:attention_modules} qualitatively and quantitatively measure their effectiveness by dropping the three attentions one by one.
Global Self Attention (GSA) has the highest impact on realism (qualitatively in Fig.~\ref{fig:attention_modules}) and diversity as expected, because this accounts for the attentions between every pair of nodes.
%
Relational Cross Attention (RCA) enforces connections between rooms and doors and has the most impact on the compatibility metric.
Component-wise Self Attention (CSA) focuses on individual rooms and reveals a significant impact on the room shape quality in Fig.~\ref{fig:attention_modules}, causing self-intersections and ``impossible" shapes without it.
The second last row of Table~\ref{tab:different_attentions} is a version with three times more GSA attention layers instead of the combination of CSA, GSA, and RCA, which demonstrates the importance of using all the attention types.


\begin{table}[tb]
\caption{Quantitative evaluation of the three attention mechanisms. The second last row triples the number of GSA layers instead of the combination of CSA, GSA, and RCA.
}
\label{tab:different_attentions}
\centering
\setlength{\tabcolsep}{3pt}
\small
\scalebox{.90}{\begin{tabular}{c|c|c|c|c}
\toprule
   CSA & GSA & RCA & Diversity $(\downarrow)$ & Compatibility $(\downarrow)$\\
    \cmidrule(lr){1-1}\cmidrule(lr){2-2} \cmidrule(lr){3-3} \cmidrule(lr){4-4} \cmidrule(lr){5-5}
           & \cmark & \cmark & 9.9\scalebox{0.8}{$\pm$0.1} & 2.9\scalebox{0.8}{$\pm$0.0} \\
    \midrule
    \cmark &  & \cmark       & 12.8\scalebox{0.8}{$\pm$0.2} & 4.0\scalebox{0.8}{$\pm$0.0} \\
    \midrule
    \cmark &   \cmark     &  & 11.4\scalebox{0.8}{$\pm$0.2} & 6.8\scalebox{0.8}{$\pm$0.0} \\
    \midrule
     &  \cmark \cmark \cmark  &  & 10.8\scalebox{0.8}{$\pm$0.2} & 6.5\scalebox{0.8}{$\pm$0.0} \\
    \midrule
    \cmark & \cmark & \cmark & \textbf{9.5}\scalebox{0.8}{$\pm$0.1} & \textbf{2.5}\scalebox{0.8}{$\pm$0.0} \\
    \bottomrule
\end{tabular}}
\end{table}



\begin{figure*}[t]
    \centering
    \includegraphics[width=\textwidth]{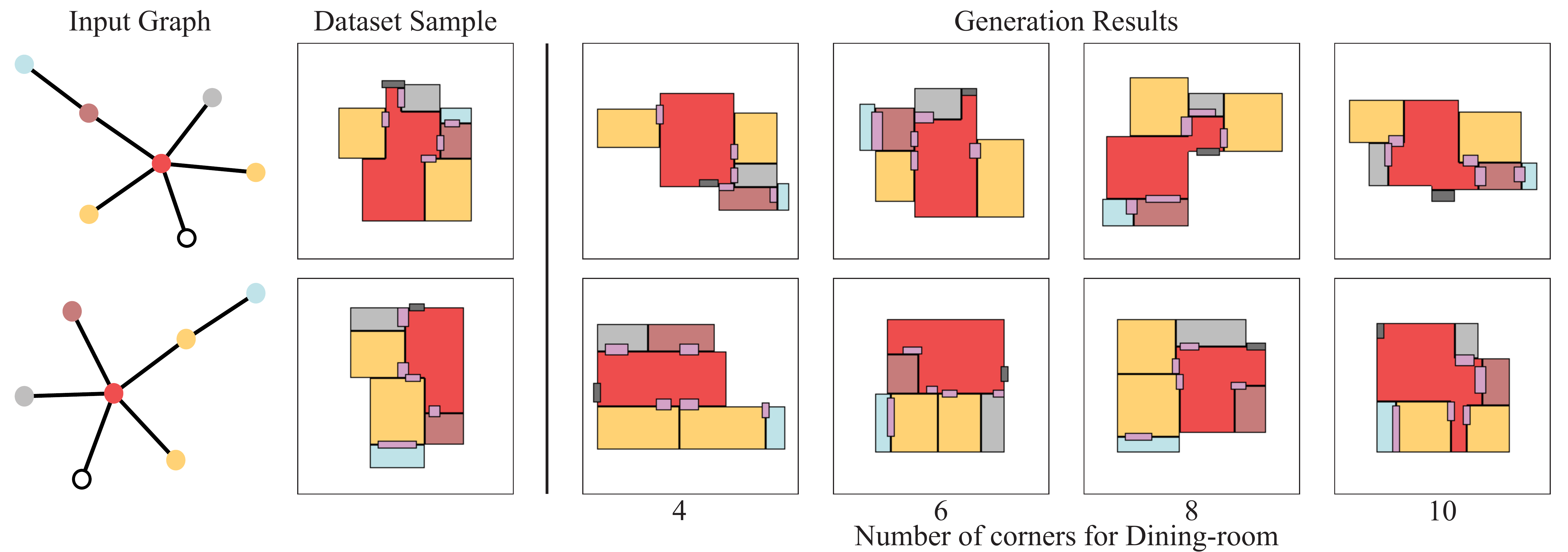}
    \caption{HouseDiffusion allows us to specify the number of corners per room.
    In this above examples, we increase the number of corners for the dining-room (red room), while keeping all the other rooms to have 4 corners.
    }
    \label{fig:manual_corner}
\end{figure*}

\mysubsubsection{Geometry modification}
Figure~\ref{fig:manual_corner} demonstrates a new capability of HouseDiffusion, allowing us to specify the exact number of corners per room,
which is not possible by a raster-based system such as House-GAN++. The figure shows that HouseDiffusion generates floorplans with increasingly more complex dining-room, where the surrounding rooms change their layouts to be consistent.

\begin{figure}[tb]
    \centering
    \includegraphics[width=\linewidth]{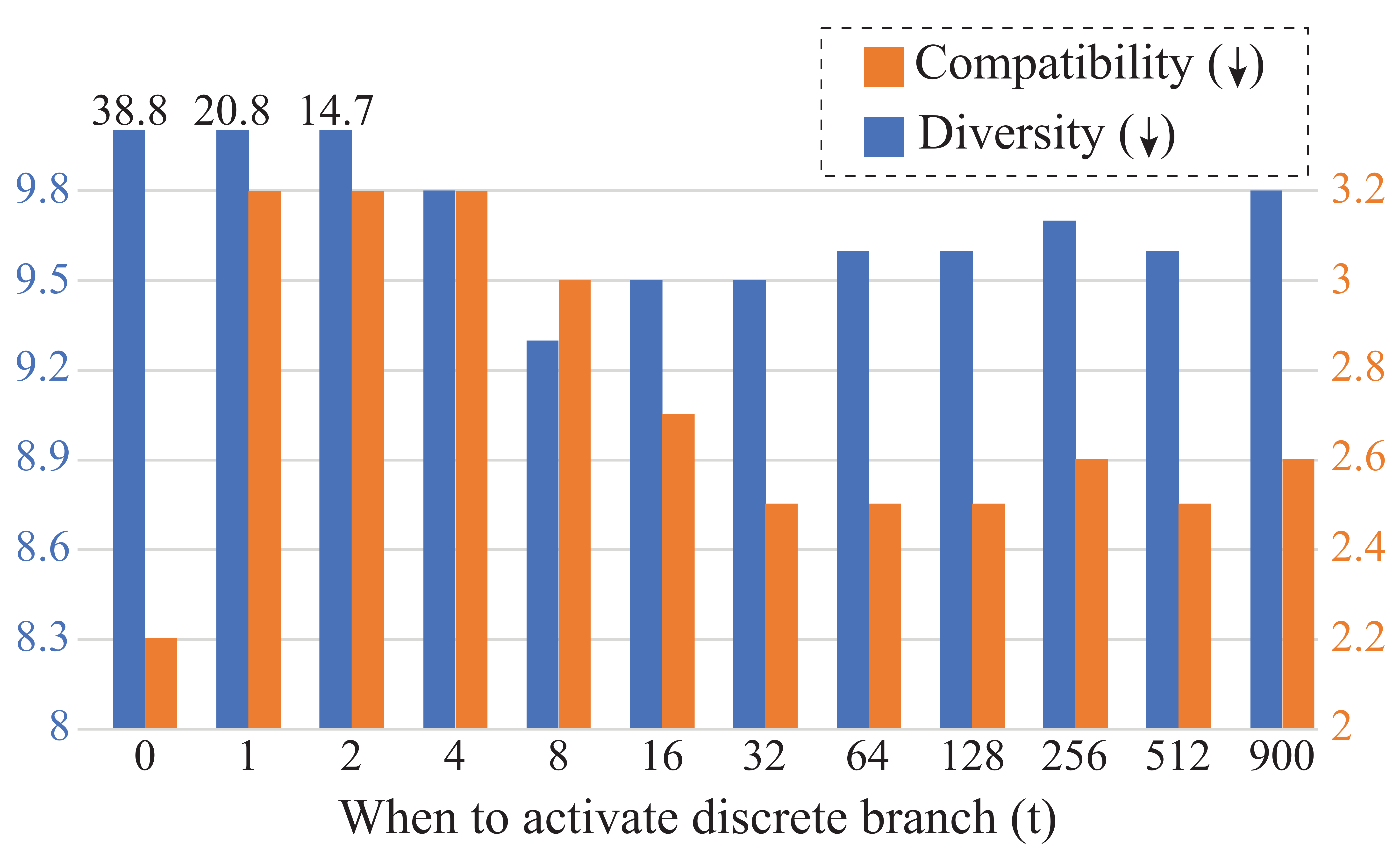}
    \caption{
    Using discrete head in a few steps gives less opportunity to the network to refine the discrete coordinates, while using it early can also lead to removing necessarily continuous informations. We fix discrete head to activate when $t<32$.}
    \label{fig:discrete_steps}
\end{figure}

\mysubsubsection{Discrete steps}
Having both discrete and continuous denoising branches is essential for generating high-quality floorplans, while
the discrete branch is active in the last 32 iterations at test time.
%
We vary this hyperparameter and measure the performance change in Fig.~\ref{fig:discrete_steps}.
Limiting to fewer iterations reduces the opportunities of discrete analysis and harms performance. On the other hand, activating the discrete branch too early also degrades the performance, because the discrete analysis is on the final floorplan shape at time 0, whose inference is not accurate at early iterations.
%
%
We found that 32 is a good number overall, which is used throughout our experiments.

\mysubsubsection{Corner augmentation}
Table~\ref{tab:input_representation} shows the corner augmentation effects in the feature embedding (\ref{eq:embedding}).
While both the diversity and compatibility improve,
the effect is higher for compatibility where the augmented points on the walls enable better analysis of geometric incident relationships.

\begin{table}[tb]
\caption{The effects of the corner augmentation in the feature embedding (\ref{eq:embedding}).}
\label{tab:input_representation}
\centering
\setlength{\tabcolsep}{3pt}
\small
\begin{tabular}{c|c|c}
\toprule
   Domain & Diversity $(\downarrow)$ & Compatibility $(\downarrow)$ \\
    \cmidrule(lr){1-1}\cmidrule(lr){2-2}\cmidrule(lr){3-3}
    Single Corner & 10.0\scalebox{0.8}{$\pm$0.3} & 3\scalebox{0.8}{$\pm$0.0} \\
    \midrule
    Augmented Corner & 9.5\scalebox{0.8}{$\pm$0.1} & 2.5\scalebox{0.8}{$\pm$0.0} \\
    \bottomrule
\end{tabular}
\end{table}

\section{Conclusion}
This paper presents a novel floorplan generative model that directly generates vector-graphics floorplans. The approach uses a Diffusion Model with a Transformer network module at the core, which denoises 2D pixel coordinates both in discrete and continuous numeric representations. The discrete representation ensures precise geometric incident relationships among rooms and doors. The transformer module has three types of attentions that exploit the structural relationships of architectural components.
Qualitative and quantitative evaluations demonstrate that the proposed system makes significant improvements over the current state-of-the-art with large margins in all the metrics, while boasting new capabilities such as the generation of non-Manhattan structures or the exact specification of the number of corners. This paper is the first compelling method to directly generate vector-graphics structured geometry.
Our future work is the handling of large-scale buildings. We will share all our code and models.


\mysubsubsection{Acknowledgement} This research is partially supported by NSERC Discovery Grants with Accelerator Supplements and DND/NSERC Discovery Grant Supplement.


{\small
\bibliographystyle{ieee_fullname}
\bibliography{egbib}
}

\end{document}